\documentclass[a4paper]{styles/svproc}
\usepackage{url}

\usepackage{graphicx}
\usepackage{algorithm}
\usepackage{algpseudocode}
\usepackage{amsmath}

\begin{document}
\mainmatter              

\title{Online Mapping for Autonomous Driving : Addressing Sensor Generalization and Dynamic Map Updates in Campus Environments}
%
\titlerunning{Online Mapping in Campus Environments}  
%
\author{Zihan Zhang \and  Abhijit Ravichandran \and
Pragnya Korti \and Luobin Wang \and Henrik I. Christensen}
\authorrunning{Zihan Zhang et al.} 
%
%
\institute{University of California, San Diego, La Jolla 92092, USA\\
\email{ziz359@ucsd.edu}}

\maketitle              

\begin{abstract}
High-definition (HD) maps are essential for autonomous driving, providing precise information such as road boundaries, lane dividers, and crosswalks to enable safe and accurate navigation. However, traditional HD map generation is labor-intensive, expensive, and difficult to maintain in dynamic environments. To overcome these challenges, we present a real-world deployment of an online mapping system on a campus golf cart platform equipped with dual front cameras and a LiDAR sensor. Our work tackles three core challenges: (1) labeling a 3D HD map for campus environment; (2) integrating and generalizing the SemVecMap~\cite{SemVecNet} model onboard; and (3) incrementally generating and updating the predicted HD map to capture environmental changes. By fine-tuning with campus-specific data, our pipeline produces accurate map predictions and supports continual updates, demonstrating its practical value in real-world autonomous driving scenarios.


\end{abstract}

\section{Introduction}
\begin{figure}[t]
    \centering
    \includegraphics[width=0.55\linewidth]{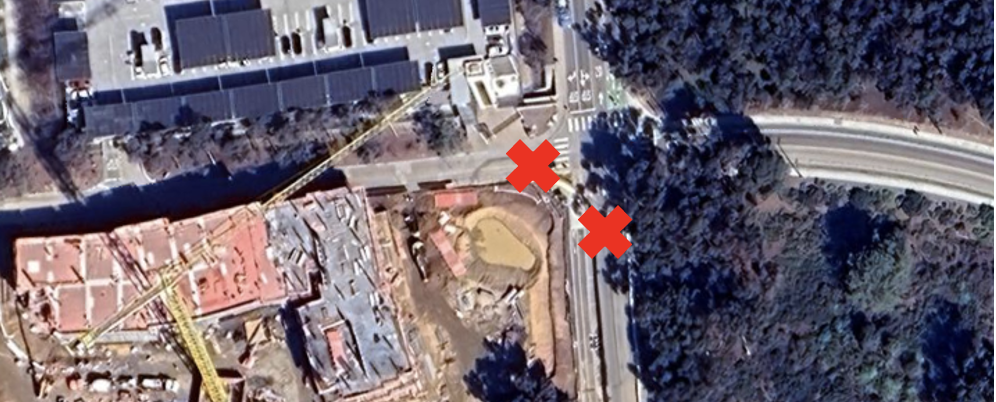}
    \caption{Campus construction zone; regions marked with red crosses indicate areas where data collection was not possible.}
    \label{construction}
\end{figure}
High-definition (HD) maps play a critical role in autonomous driving by supplying precise geometric and semantic information—such as road boundaries, lane dividers, and crosswalks—to enable accurate localization, prediction, and planning~\cite{DES}.
However, producing HD maps remains costly and labor-intensive. Practitioners must manually annotate centimeter-level road geometry, sequence those details into waypoints, and rigorously validate the result. Maintaining such maps is even harder, because the physical world changes constantly and reliably detecting every local modification is still an open problem.
This poses a significant barrier to scalability, especially in dynamic or previously unmapped environments. For example, Fig.\ref{construction} shows an area under construction in our campus environment, where the roads marked by a red cross were undrivable before, and we were unable to obtain the ground truth map for that area.

Recent progress in perception and learning-based systems has enabled a new paradigm: online mapping, where map elements are automatically generated from onboard sensors such as cameras and LiDAR. Despite promising results in static benchmarks \cite{nuScenes}, these models are rarely evaluated in real-world systems, where sensor generalization and environmental changes pose challenges. 

In this work, we provide a step forward in closing the gap between online mapping research and real-world deployment. We present a complete real-world deployment of an online mapping system \cite{SemVecNet} on a campus autonomous driving platform equipped with dual front-facing cameras and a LiDAR sensor \cite{golfcart}. Additionally, we propose a clustering-based method to construct HD map priors from online mapping results. Our system enables the generation and continual update of HD maps in previously unseen environments. Our contributions are as follows:
\begin{itemize}

    \item \textbf{Real-world deployment:} We demonstrate the feasibility of using a sensor-generalizable online mapping model in a real-world autonomous driving system, addressing the challenges of sensor variation and environmental dynamics
    
    \item \textbf{3D ground truth map labeling:} We present our system of generating fine-grained HD annotations, enabling high-quality supervision in unseen environments.
    
    \item \textbf{Model fine-tuning and adaptation:} We fine-tune a mapping model SemVecNet~\cite{SemVecNet} using campus-specific labels to enhance performance in unseen environments, showing significant improvements across varied road scenarios.
    
    \item \textbf{Incremental map update mechanism:} We develop a novel pipeline that accumulates multi-frame map predictions to incrementally refine the HD map, enabling the generation new map and detection of changes such as construction and reconfigured lanes.
\end{itemize}



\section{Related Work}
Online mapping models~\cite{hdmapnet,vectormapnet,Streammapnet,maptr,SemVecNet} present a promising solution for autonomous driving by dynamically generating map geometry and semantics from sensor inputs. These models aim to reduce the reliance on manual annotations and enable adaptive mapping in changing environments. For instance, HDMapNet~\cite{hdmapnet} and VectorMapNet~\cite{vectormapnet} utilize multi-camera and LiDAR data to produce vectorized HD map elements, while MapTR~\cite{maptr} and StreamMapNet~\cite{Streammapnet} rely solely on monocular images, enhancing deployment flexibility. However, most of these models are trained and evaluated on fixed datasets and exhibit strong dependence on the specific sensor configurations and environmental characteristics seen during training. As a result, they often suffer from overfitting and show degraded performance when transferred to new platforms or unseen domains. SemVecNet~\cite{SemVecNet} introduces a two-stage framework that first generates an intermediate semantic map and then converts it into a vectorized representation, improving robustness to cross-dataset variation. While this approach achieves better generalization, its performance still falls short of the reliability needed for real-world deployment in dynamic and diverse environments.

\section{Technical Approach}

\begin{figure}[t]
    \centering
    \includegraphics[width=1\linewidth]{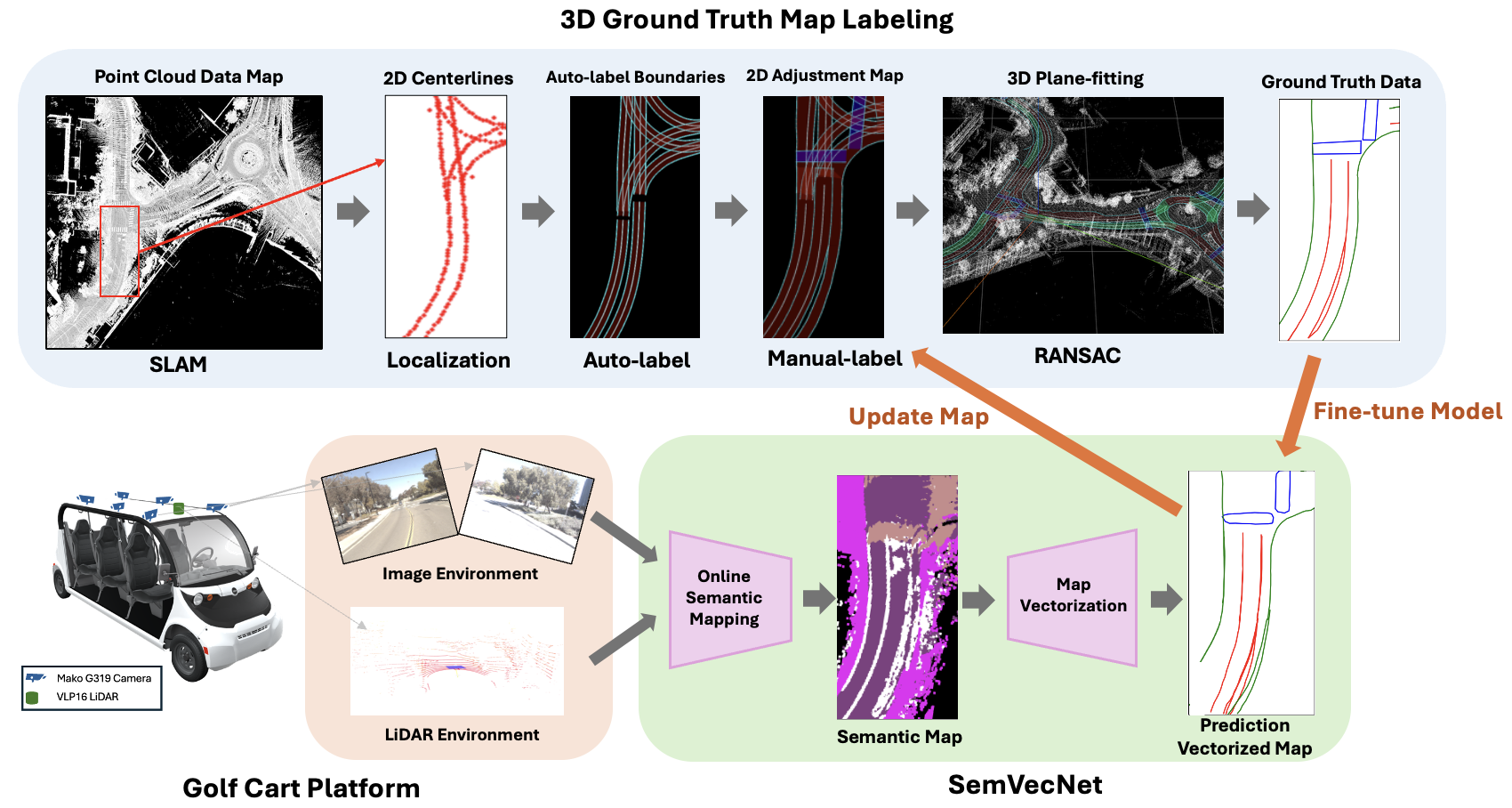}    \caption{Overview: The blue section represents the 3D ground truth map labeling pipeline, the orange section denotes the golf cart platform, and the green section corresponds to the online mapping SemVecNet\cite{SemVecNet} pipeline. The orange arrows illustrate the interaction steps between the map labeling and online mapping pipelines for model fine-tuning and map updates.}
    \label{overview}
\end{figure}

Our approach addresses two key challenges in real-world autonomous driving: sensor generalization and dynamic HD map maintenance. An overview of the proposed pipeline is shown in Fig.\ref{overview}. The system is composed of three main components.

\subsection{3D Ground Truth Map Labeling} \label{Ground Truth Map Generation}
To create an accurate and high-resolution HD map for our autonomous golf cart system, we developed a multi-step pipeline, illustrated in the blue section of Fig.~\ref{overview}.
\begin{itemize}
    \item \textbf{3D Point Cloud (PCD) Map:} We use Normal Distribution Transform (NDT)\cite{NDT} scan matching—a SLAM (Simultaneous Localization and Mapping) technique—on Velodyne LiDAR data to incrementally build a dense 3D point cloud map. This map provides a structural foundation for HD map construction, capturing detailed road and environmental features.
    \item \textbf{Centerline Extraction:} Leveraging SLAM-based localization, we extract lane centerlines by tracing the vehicle’s $(x, y)$ positions along drivable paths. These centerlines define the map’s geometric backbone and guide subsequent labeling.
    \item \textbf{2D Auto-labeling:} Around each centerline, we automatically generate lane boundaries by extending 10 feet to each side, following California Highway Design Manual standards~\cite{highway}. This yields a preliminary 2D vector map in Lanelet2 format.
    \item \textbf{2D Manual Adjustment:} Using Autoware Tools~\cite{MapTool}, we refine the 2D map by aligning it with the point cloud and annotating features such as crosswalks and stop signs. This step ensures accuracy and completeness in real-world lane representation.
    \item \textbf{3D Plane Fitting:} To recover elevation, we fit road surfaces using RANSAC~\cite{RANSAC} and assign height values to each 2D lane point using k-d tree search~\cite{kdtree} over inlier ground points. This results in a complete 3D HD map aligned with the physical road surface.
\end{itemize}

\subsection{SemVecMap Model Integration and Fine-Tuning}
 Following the SemVecNet\cite{SemVecNet} architecture,  we address the challenge of sensor generalization by conditioning the model on an intermediate semantic map $S_t$, rather than directly predicting vectorized outputs from raw camera data. This semantic conditioning allows the model to adapt more effectively to different sensor configurations and environments. 
 In our setup, the online mapping task processes two synchronized front-facing camera images, $I_t = \{I_1, I_2\}$, along with the corresponding LiDAR point cloud $L_t$. The first stage of the model generates the semantic map $S_t$ from these inputs. The semantic map then serves as a conditioning input for the second stage, which outputs the vectorized map $M_t$.
 The final output $M_t$ is composed of a set of $n$ point sequences, $P_t = \{P_1, P_2, \dots, P_n\}$, where each $P_i$ represents a vectorized map element, such as a lane boundary, a road divider or a pedestrian crosswalk. This modular pipeline enables more robust and interpretable mapping, while also facilitating fine-tuning on our campus dataset.
\begin{enumerate}
    \item \textbf{Online Semantic Mapping Generation:} We use MScale-HRNet\cite{HRNet} to generate semantic segmentations for each image in $I_t$. To enable efficient deployment, the model is optimized using TensorRT~\cite{TensorRT} to reduce inference latency. Once the semantic segmentation outputs are obtained, the LiDAR point cloud $L_t$ is projected onto the segmented images, associating depth information to produce a semantic point cloud. The final semantic map $S_t$ is then constructed by selecting the class with the highest probability at each point.
    \item \textbf{Map Vectorization:}  The semantic map $S_t$ is first processed by a ResNet-50~\cite{ResNet-50} encoder to extract bird's-eye view (BEV) feature representations, denoted as $F_t$. These features are then passed to a transformer-based decoder, which operates over a structured set of learnable queries. The hierarchical query for the $j$-th point of the $i$-th map element is formulated as 
    \begin{equation}
        q^{\text{hie}}_{ij} = q^{\text{ins}}_i + q^{\text{pt}}_j
    \end{equation}
    which ${q^{\text{ins}}_i}$ indicates a set of instance-level queries and  ${q^{\text{pt}}_j}$ indecates a  set of point-level queries. The decoder applies decoupled self-attention across the map queries, enabling information exchange between instance-level and point-level representations. This is followed by deformable cross-attention between the hierarchical queries and the BEV feature map $F_t$, allowing the model to effectively incorporate spatial context from the visual scene.
    The decoder outputs two prediction heads: a classification branch that estimates instance-level scores, and a regression branch that predicts 2D coordinates for each point in the fixed-size point set. Each output $P_i$ contains 20 points, representing structured vectorized elements such as boundaries, dividers, or crosswalks.

    \item \textbf{Fine-Tuning:} To adapt the model to the unique characteristics of our campus environment, we fine-tune the map vectorization module using our ground-truth annotated dataset. Consistent with the SemVecNet~\cite{SemVecNet} training procedure, we apply focal loss~\cite{focalloss} for instance classification, a point-to-point loss for accurate vertex regression, and an edge direction loss to preserve geometric consistency between adjacent points.

\end{enumerate}

\begin{algorithm}[t]
\caption{Clustering Consistent Line Structures from Frame-by-Frame Predictions}
\label{map_update_algorithm}
\begin{algorithmic}
\State Initialize grid \texttt{accumulation\_grid[$W \times H$]} = 0
\State Initialize empty set \texttt{new\_map\_elements}
\State Initialize empty set \texttt{update\_map\_elements}
\For{each frame in \texttt{frames}}
    \For{each line in frame}
        \For{each point $(x, y)$ along the line}
            \State Convert ego-frame $(x, y)$ to map-frame $(x', y')$
            \State Convert $(x', y')$ to grid indices $(i, j)$
            \State \texttt{accumulation\_grid[$i$, $j$]} += 1
        \EndFor
    \EndFor
\EndFor
\State \texttt{dense\_mask} = \texttt{accumulation\_grid} $>$ \texttt{threshold}
\State \texttt{extract\_lines} = \texttt{skeletonize}(\texttt{dense\_mask})

\For{each extract line in \texttt{extract\_lines}}
    \State Convert grid indices of extract line to map-frame coordinates
    \State \texttt{new\_map\_elements$.$add(result)}
\EndFor

\For{each grid\_cell (size 30m $\times$ 30m) in map }
    \State Compute mAP between \texttt{new\_map\_elements} and \texttt{existing\_HD\_map}
    \If{mAP $<$ update threshold}
        \State \texttt{update\_map\_elements$.$add(grid\_cell)}
    \EndIf
\EndFor
\State \Return \texttt{update\_map\_elements}
\end{algorithmic}
\end{algorithm}
\subsection{Semi-Auto Map Update}

To address the challenge of maintaining up-to-date HD maps in dynamic environments, we propose a semi-automated map update mechanism. As the golf cart navigates the campus, the fine-tuned SemVecMap model continuously generates predictions of map elements in an ego-centric frame, covering a $(30 \times 60)$ meter region. These predicted map elements, once transformed into global map coordinates, often contain overlapping segments due to repeated observations. To handle this redundancy, we construct an accumulation grid with a resolution of 0.5 meters, aggregating overlapping predictions to generate a dense confidence mask. Line structures are then extracted from the skeleton through thresholding and filtering. To identify regions requiring updates, we partition the entire map into $30 \times 30$ meter grid cells and compute the Average Precision (AP) metric between the newly predicted elements and the existing HD map within each cell. Low-AP regions are flagged as candidates for update, indicating either structural changes or previously unlabeled areas. Once such regions are identified, we employ Autoware Tools to overlay the new predictions onto the existing HD map. This process enables conflict resolution—e.g., handling mismatches between historical and current data—and allows for selective updates that reflect the most recent environmental conditions, while preserving reliable segments of the prior map. The complete procedure is summarized in Algorithm~\ref{map_update_algorithm}.

\section{Experiments}
\subsection{Hardware}
Our autonomous driving platform~\cite{golfcart} is built on a six-seat GEM e6 golf cart retrofitted with a drive-by-wire (DBW) system, onboard computing units, and a full sensor suite for perception and mapping. The vehicle is equipped with multiple Mako G319 cameras and a Velodyne VLP-16 LiDAR, as shown in Fig.~\ref{overview} (bottom left). Camera intrinsics are calibrated using a checkerboard, and extrinsic calibration between cameras and LiDAR is performed via manual correspondence. Accurate calibration is essential for reliable online map generation. The onboard system includes two NVIDIA RTX 3070 GPUs, enabling real-time execution of the online mapping pipeline alongside other learning-based modules.

\subsection{Campus Dataset Collection}
\begin{table}[t]
\centering
\caption{Distribution of Training and Testing Scenes by Scenario Type}
\label{scene_distribution}
\begin{tabular}{|c|c|c|}
\hline
\textbf{Scenario Type} & \textbf{Training Scenes} & \textbf{Testing Scenes} \\
\hline
Straight Road & 843 & 282 \\
Intersection  & 429 & 144 \\
Loop          & 179 & 60  \\
Roundabout    & 114 & 39  \\
Multi-lane    & 18  & 6   \\
\hline
\textbf{Total} & \textbf{1583} & \textbf{531} \\
\hline
\end{tabular}
\end{table}
\begin{figure}[t]
    \centering
    \includegraphics[width=0.6\linewidth]{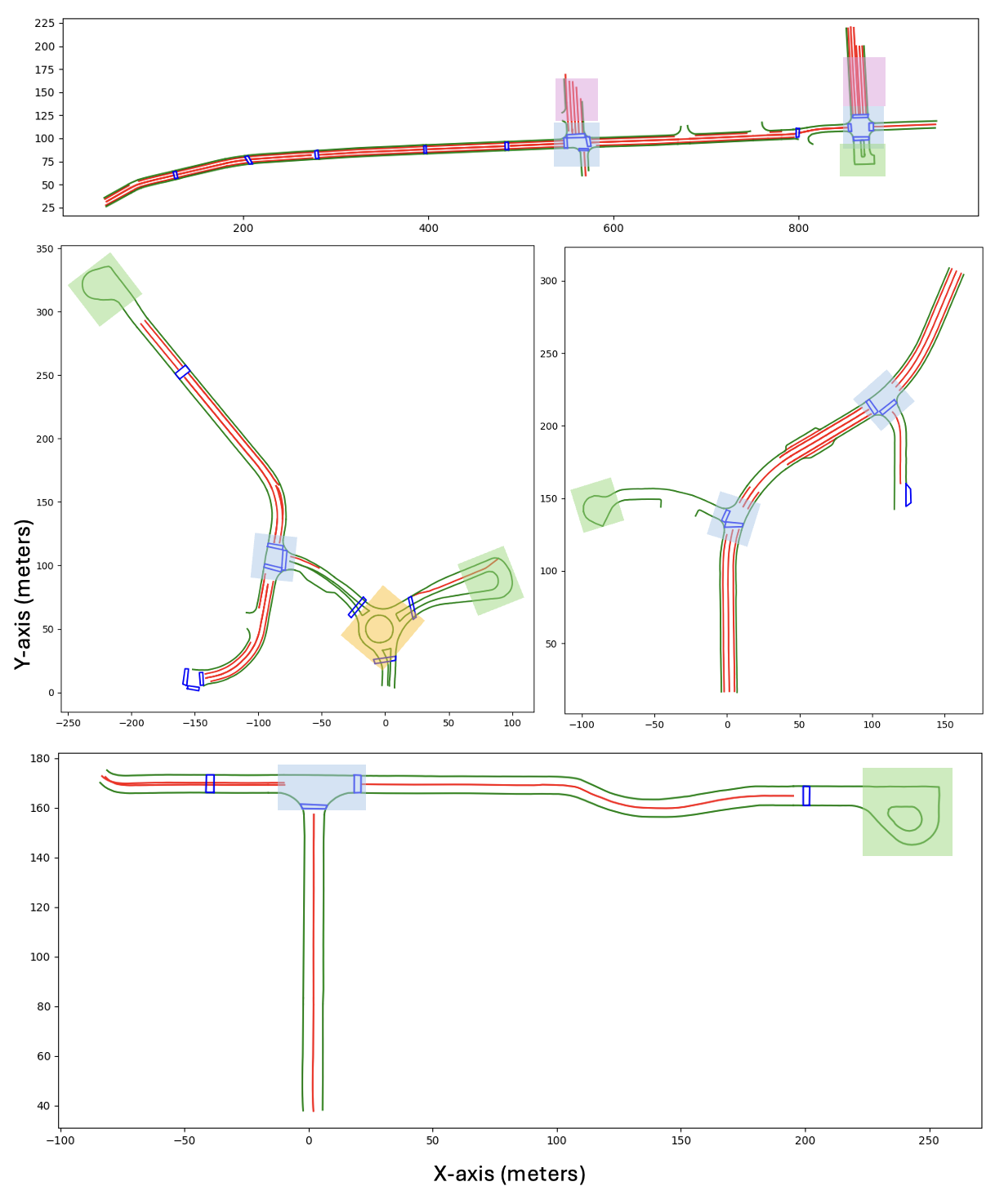}
    \caption{Four main annotated regions in the campus map. Green polylines denote road boundaries, red polylines represent lane dividers, and blue polygons indicate pedestrian crosswalks. Key regions of interest are highlighted—yellow represents the roundabout, blue indicates the intersection, green indicates the loop, and pink denotes the multi-lane two-way road.}  
    \label{fourarea}
\end{figure}

We collected and labeled ground truth data from four designated areas on campus using the technical pipeline described in Section~\ref{Ground Truth Map Generation}. These regions span approximately 3,300 meters (one-way) and include a diverse set of lane structures: two-way straight roads, intersections (blue area), roundabouts (yellow area), loops (green area), and multi-lane two-way roads (pink area), as shown in Fig.~\ref{fourarea}. The inclusion of such varied road geometries provides a rich dataset for model fine-tuning, allowing the system to generalize across different driving scenarios present in the campus environment.

During data collection, the golf cart was manually driven through these four areas while recording data from the main camera and LiDAR sensors. Data was initially captured at 10 Hz. To mitigate redundancy and reduce the risk of overfitting during fine-tuning, the dataset was subsequently downsampled to 2 Hz. Each recorded segment was annotated according to the scenario type—straight road, intersection, loop, roundabout, or multi-lane-and the dataset was random split into 75\% for training and 25\% for testing base on different scenario type. Table~\ref{scene_distribution} summarizes the number of scenes collected for each scenario type.

\subsection{Metrics}
We use Average Precision (AP) metric for evaluation our fine-tuning performance and also the comparision between the new prediction map and old map.
The AP is calculated separately for each class. The final AP score is computed by averaging the AP values across three Chamfer distance thresholds $\theta \in \{0.5\,\text{m}, 1.0\,\text{m}, 1.5\,\text{m}\}$ and the number of classes to obtain the mean AP (mAP).

The Chamfer distance between two point sets $A$ and $B$ in the same perception range is defined as:

\begin{equation}
\mathrm{D}_{\text{Chamfer}}(A, B) = \frac{1}{|A|} \sum_{a \in A} \min_{b \in B} \|a - b\|_2^2 + \frac{1}{|B|} \sum_{b \in B} \min_{a \in A} \|b - a\|_2^2
\end{equation}
\subsection{Quantitative Performance}
The results in Table~\ref{finetune_performance} reveal a clear domain gap between autonomous driving datasets, particularly in cross-dataset generalization. SemVecNet\cite{SemVecNet} model trained on NuScenes\cite{nuScenes} performs well on its own validation set (mAP of 48.8) but drops sharply when evaluated on Argoverse 2\cite{argoverse2} (14.8) and Campus (13.0), highlighting the limitations of transferring the model across different sensor setups, road geometries, and annotation standards. Incorporating domain-specific data significantly improves performance. When Campus data is added during training, the mAP on the Campus test set rises from 13.0 to 56.9, demonstrating the value of environment-specific fine-tuning for real-world deployments.Table~\ref{scenario_performance} presents an ablation study across various road scenarios. In all cases, fine-tuning yields substantial gains. For example, in loop structures—often underrepresented in standard datasets—the mAP increases from 5.82 to 47.51. This indicates that scenario diversity introduces varying levels of difficulty, and that fine-tuning enables better generalization to complex geometries. Overall, these results underscore the challenges of cross-dataset deployment and the critical role of targeted fine-tuning in achieving reliable and accurate map generation in dynamic, real-world environments.
\begin{table}[t]
\centering
\caption{Fine-tuning Evaluation Performance Across Datasets}
\label{finetune_performance}
\begin{tabular}{|c|c|c|c|c|c|}
\hline
\textbf{Training Dataset} & \textbf{Testing Dataset} & \textbf{AP$_\mathrm{boundary}$} & \textbf{AP$_\mathrm{divider}$} & \textbf{AP$_\mathrm{crosswalk}$} & \textbf{mAP} \\
\hline
NuScenes & Argoverse 2 & 27.8 & 13.1 & 3.5 & 14.8 \\
NuScenes & NuScenes    & 54.5 & 50.5 & 41.4 & 48.8 \\
\hline
NuScenes & Campus       & 13.1 & 13.5 & 12.4 & 13.0 \\
NuScenes + Campus & Campus & 65.9 & 50.3 & 54.5 & 56.9 \\
\hline
\end{tabular}
\end{table}

\begin{table}[t]
\centering
\caption{Performance by Scenario Type. Rows marked with  \textbf{*} indicate results before fine-tuning.}
\label{scenario_performance}
\begin{tabular}{|c|c|c|c|c|}
\hline
\textbf{Scenario Type} & \textbf{AP$_\mathrm{boundary}$} & \textbf{AP$_\mathrm{divider}$} & \textbf{AP$_\mathrm{crosswalk}$} & \textbf{mAP} \\
\hline
Straight Road   & 78.88 & 52.29 & 49.07 & 60.08 \\
 \textbf{*} & 16.40 & 14.54 & 5.30 & 12.08 \\
\hline
Intersection    & 61.98 & 46.43 & 59.86 & 56.09 \\
 \textbf{*} & 13.43 & 12.69 & 17.09 & 14.40 \\
\hline
Loop            & 56.85 & 24.98 & 60.68 & 47.51 \\
 \textbf{*} & 3.97  & 8.41  & 5.07  & 5.82  \\
\hline
Roundabout      & 40.49 & 51.87 & 31.85 & 41.40 \\
 \textbf{*} & 9.76  & 4.32  & 0.99  & 5.02  \\
\hline
Multi-lane      & 90.53 & 91.67 & 81.48 & 87.89 \\
 \textbf{*}  & 15.50  & 18.01  & 11.11  & 14.87  \\
\hline
\end{tabular}
\end{table}

\subsection{Qualitative Visualization}
\begin{figure}[t]
    \centering
    \includegraphics[width=1\linewidth]{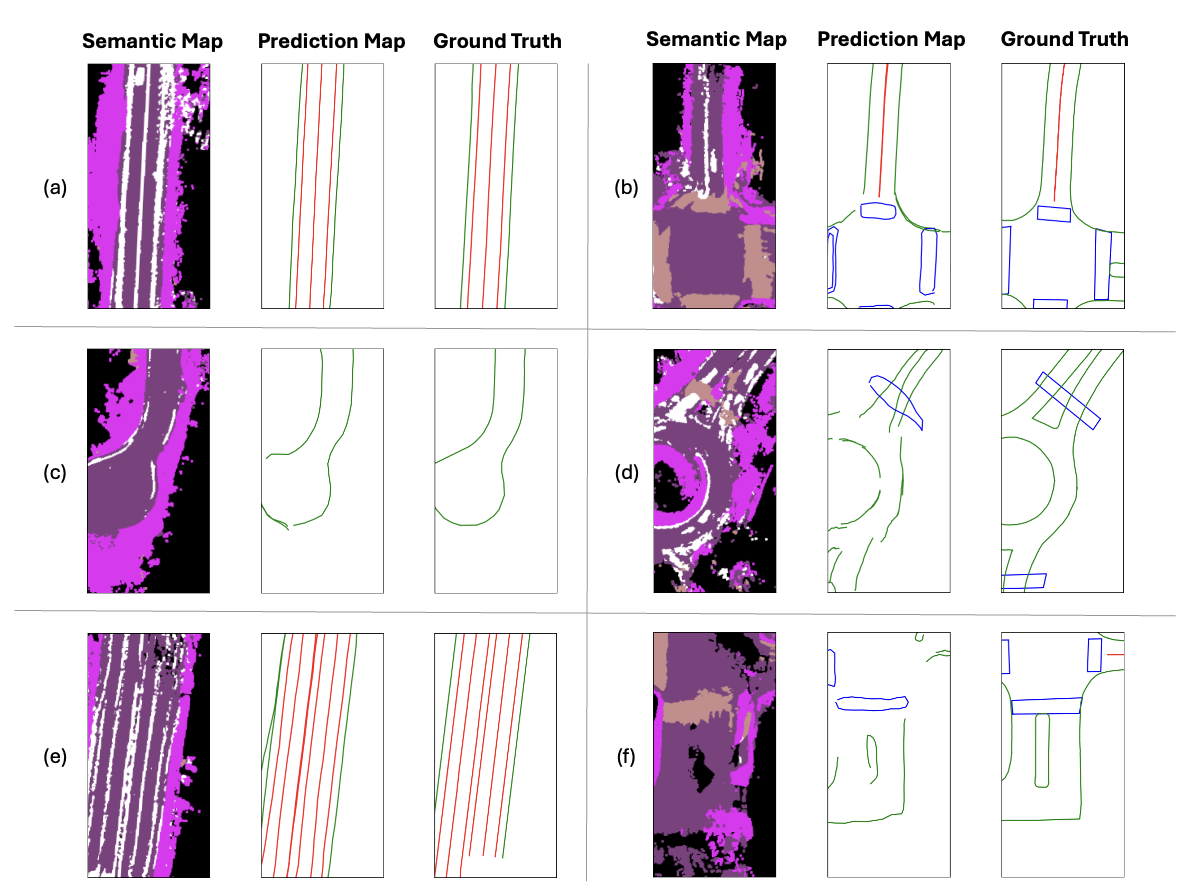}
    \caption{Qualitative visualization of the fine-tuned online mapping model across various driving scenarios in the campus environment. Each triplet shows the semantic map input (left), the predicted vectorized map (middle), and the ground truth annotation (right). The examples include straight roads (a)(e), intersections(b)(f), loop(c), and roundaboud(d).}
    \label{visualization_performance}
\end{figure}
In this section, we presents six sets of visualizations that qualitatively demonstrate the performance of our fine-tuned online mapping model across diverse campus driving scenarios in Figure \ref{visualization_performance}. Each triplet consists of the input semantic map (left), the predicted vectorized map (middle), and the ground truth (right). The examples include various road configurations: straight roads (a, e), intersections (b, f), a loop (c), and roundabouts (d). These visualizations highlight the model's ability to extract structured vector representations from semantic observations. In scenarios such as (a) and (e), the model successfully reconstructs dense lane divider and boundary lines in straight, multi-lane roads. Intersections and roundabouts shown in (b) and (d) are also well handled, with the predicted lane and crosswalk aligning closely with the ground truth. Example (c) demonstrates accurate modeling of the loop boundary, further showcasing generalization to non-linear topologies. A failure case is shown in (f), where heavy noise and occlusion in the semantic input lead to distorted predictions. This underscores the importance of temporal aggregation across frames, rather than relying on single-frame observations, to ensure robust map updates in dynamic or partially observed environments.

\subsection{Map Generation Case}
\begin{figure}[t]
    \centering
    \includegraphics[width=0.95\linewidth]{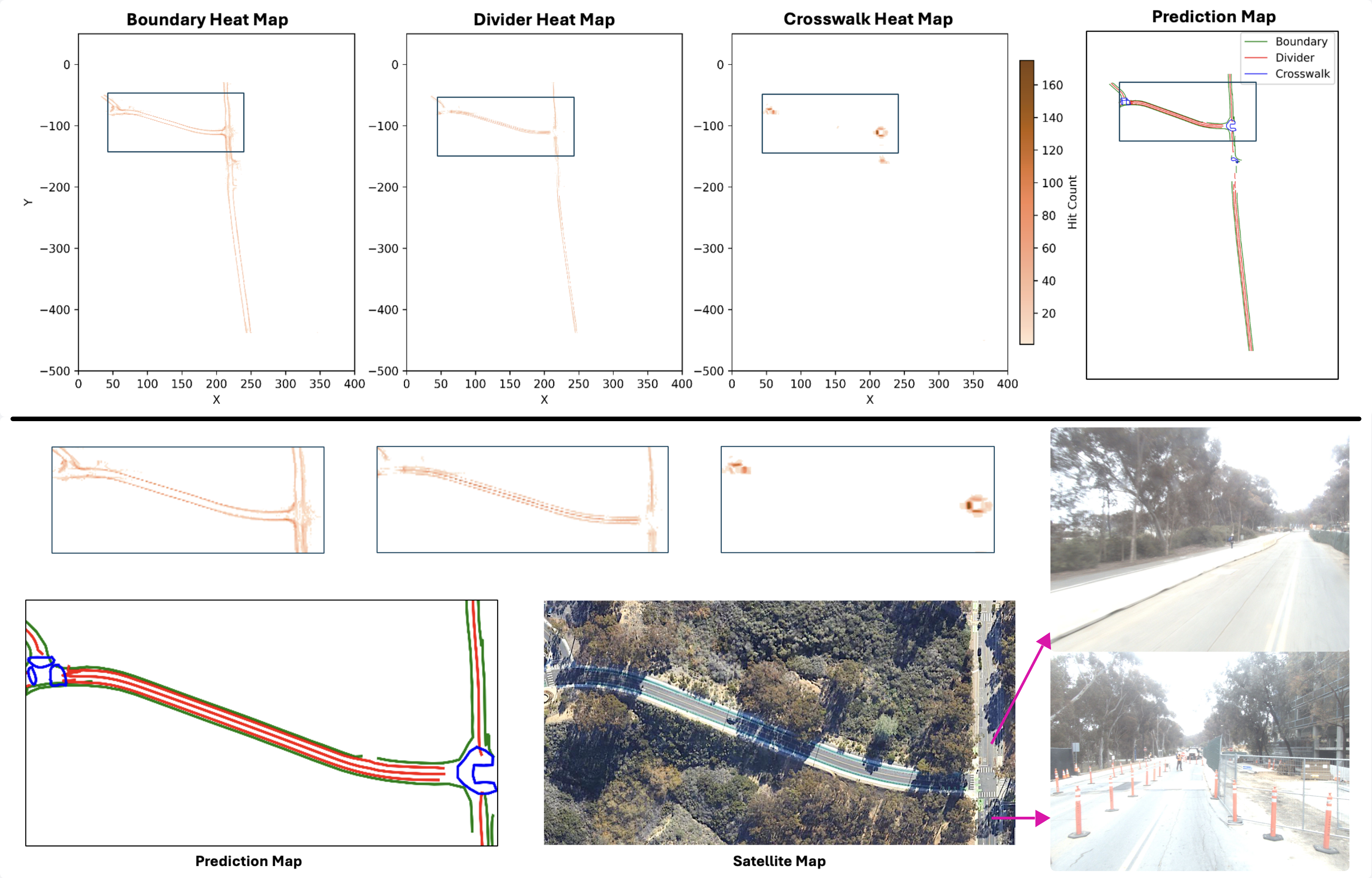}
    \caption{New map generation in unseen areas using our fine-tuned online mapping model. Top: heatmaps of boundaries, dividers, and crosswalks accumulated from multiple frames, with the final vectorized map extracted from high-confidence regions. Bottom: zoom-in views of two intersections, comparing the prediction map with the old satellite map and the current camera views.}

    \label{Generate_map}
\end{figure}
We evaluate our fine-tuned online mapping model and update algorithm by generating a vectorized map for a previously unseen area—excluded from both training and testing datasets. As shown in Figure~\ref{Generate_map}, the top row presents heatmaps for three predicted classes (boundary, divider, and crosswalk) aggregated from sequential frames. These heatmaps capture detection density and temporal consistency. The final prediction (top-right) is derived from high-confidence regions in these maps. The lower portion of the figure zooms into the highlighted region, focusing on two intersections. The predicted map (bottom-left) is compared against satellite imagery (middle) and a real-world camera view (right). The model demonstrates high accuracy, successfully capturing lane dividers, crosswalks, and even subtle features like bicycle lanes. A particularly notable case is the intersection on the right, which is under construction. Unlike the old satellite map, the predicted layout correctly reflects the narrowed upper road segment, blocked off by temporary white barriers. However, the lower section is occluded by orange traffic delineators, resulting in a noisy road geometry that the model hard captures. This case study highlights the system’s ability to generate accurate, real-time maps in unseen environments and adapt to on-the-fly changes—an essential capability for scalable autonomous navigation.

\section{Conclusions}
In this paper, we presented a real-world deployment of an online mapping system integrated into a campus autonomous vehicle platform. Our system leverages onboard sensors—including cameras and LiDAR—to generate high-definition maps. To support effective fine-tuning and evaluation, we also introduced a 3D labeling pipeline for generating high-quality ground truth annotations in previously unmapped environments. Through extensive experiments across diverse campus scenarios, we demonstrated that fine-tuning with environment-specific data significantly improves mapping accuracy and generalization. Additionally, our incremental update strategy enables the system to generate a new map or adapt to dynamic changes, such as altered road layouts. Our results validate the feasibility and effectiveness of sensor-driven, continuously updating HD map generation in real-world autonomous driving applications. This work highlights the practical potential of online mapping pipelines and offers a foundation for scalable map maintenance beyond static datasets.

\bibliography{paper}{}

\end{document}